%% file: acl_latex.tex
\documentclass[11pt]{article}

\usepackage{acl}
\usepackage{titletoc}
\usepackage{times}
\usepackage{latexsym}
\usepackage{booktabs}
\usepackage{multirow}
\usepackage{subcaption} 
\usepackage[T1]{fontenc}

\usepackage[utf8]{inputenc}

\usepackage{microtype}

\usepackage{inconsolata}

\usepackage{graphicx, amsmath, amssymb, amsthm, bm, algorithm, algorithmic}

\title{GIFT: Reconciling Post-Training Objectives via Variational Finite-Temperature Gibbs Initialization}

\author{
  Zhengyang Zhao$^{1,*}$, 
  Lu Ma$^{1,*}$, 
  Yizhen Jiang$^{1,2}$, 
  Xiaochen Ma$^1$, 
  Zimo Meng$^1$, \\
  \textbf{Chengyu Shen}$^1$, 
  \textbf{Lexiang Tang}$^1$, 
  \textbf{Haoze Sun}$^2$, 
  \textbf{Peng Pei}$^2$, 
  \textbf{Wentao Zhang}$^{1,3,4,\dagger}$ \\
  $^1$Peking University \quad 
  $^2$Meituan \quad
  $^3$Zhongguancun Academy \\
  $^4$Beijing Key Laboratory of Data Intelligence and Security (Peking University) \\
  \texttt{zhengyangzhao25@stu.pku.edu.cn}, \texttt{maluqaq@163.com} 
}

\begin{document}
\maketitle

\renewcommand{\thefootnote}{\fnsymbol{footnote}}
\footnotetext[1]{Equal contribution.}
\footnotetext[2]{Corresponding author.}
\renewcommand{\thefootnote}{\arabic{footnote}} 

\begin{abstract}
The prevailing post-training paradigm for Large Reasoning Models (LRMs)—Supervised Fine-Tuning (SFT) followed by Reinforcement Learning (RL)—suffers from an intrinsic optimization mismatch: the rigid likelihood maximization in SFT induces distributional collapse, thereby exhausting the exploration space necessary for subsequent RL. Motivated by the Gibbs optimum of KL-regularized RL, we derive a token-level variational surrogate that makes the SFT target structurally compatible with the subsequent RL stage, and propose Gibbs Initialization with Finite Temperature (GIFT). Standard SFT emerges as a degenerate zero-temperature limit of this surrogate, while a finite temperature preserves structural diversity. Our experiments demonstrate that GIFT outperforms standard SFT and surpasses competitive baselines when utilized for RL initialization. Our code is available at \url{https://github.com/zzy1127/GIFT}.
\end{abstract}

\section{Introduction}
The training pipeline for Large Reasoning Models (LRMs) has converged into a standard two-stage paradigm. First, Supervised Fine-Tuning (SFT) establishes fundamental reasoning capabilities by leveraging expert demonstrations~\cite{chung2022scalinginstructionfinetunedlanguagemodels, ouyang2022traininglanguagemodelsfollow, touvron2023llama2openfoundation}. Second, Reinforcement Learning (RL) encourages the model to explore self-generated reasoning paths, allowing it to optimize for correctness beyond the constraints of static datasets~\cite{shao2024deepseekmathpushinglimitsmathematical, yu2025dapoopensourcellmreinforcement, rafailov2024directpreferenceoptimizationlanguage, schulman2017proximal}. This hierarchical approach underpins current state-of-the-art reasoning systems~\cite{deepseekai2025deepseekv3technicalreport, yang2025qwen3technicalreport, nvidia2024nemotron4340btechnicalreport,huang2026step}.

However, a fundamental theoretical discrepancy persists between these two phases. While SFT typically minimizes cross-entropy loss to imitate expert tokens, RL maximizes expected rewards through environmental exploration. Standard SFT aggressively suppresses the probability mass of non-target tokens, thereby forcing the policy distribution to collapse onto deterministic data points~\cite{wu2025generalizationsftreinforcementlearning}. This phenomenon, characterized as distributional collapse, erodes the structural priors inherited from pre-training and severely restricts the subsequent exploration space~\cite{li2025loglikelihoodprobabilitybasedobjectives}. Consequently, as the model transitions to the RL stage, it encounters a critical bottleneck: the stochastic diversity required to discover high-reward trajectories has been substantially reduced by the rigid supervision of the preceding SFT phase~\cite{gudibande2023falsepromiseimitatingproprietary, kang2025quagmiressftrlposttraininghigh}.

Previous research has attempted to mitigate this issue by heuristically tuning data mixtures and inter-stage scheduling~\cite{kang2025quagmiressftrlposttraininghigh, liu2025acereasonnemotron11advancingmath}, or by modifying the SFT objective to implicitly incorporate RL properties~\cite{zhu2025proximalsupervisedfinetuning, wu2025generalizationsftreinforcementlearning, du2025simplifyrlhfrewardweightedsft, rafailov2024directpreferenceoptimizationlanguage}. In contrast to these methods, we do not attempt to bypass the RL stage or rely on heuristic adjustments; instead, we aim to establish a mathematically principled coupling between the two phases. We posit that SFT should not be treated as an isolated imitation task but rather as a theoretically motivated initialization for subsequent RL within a coupled post-training framework.

Based on this insight, motivated by the Gibbs optimum of KL-regularized RL, we derive a token-level variational surrogate that makes the SFT target structurally compatible with the subsequent RL stage, and propose \textbf{G}ibbs \textbf{I}nitialization with \textbf{F}inite \textbf{T}emperature (\textbf{GIFT}). GIFT ensures that while the model prioritizes expert trajectories, it preserves the structural diversity of the base model, offering a finite-temperature solution to the alignment-exploration trade-off. Within our framework, standard SFT is characterized as a degenerate zero-temperature limit that induces a rigid distributional collapse onto the expert data. In contrast, GIFT maintains a finite temperature, incorporating expert supervision as a continuous energy potential rather than a hard overwrite, thereby establishing a principled distributional bridge to subsequent RL.

Empirically, we conduct SFT followed by RL on a subset of the DeepMath-103k dataset~\cite{he2025deepmath}. Experimental results demonstrate that GIFT consistently outperforms standard SFT and matches or surpasses other robust SFT variants when utilized as an initialization policy for RL across diverse mathematical reasoning benchmarks and out-of-distribution (OOD) tasks. Furthermore, our geometric and distributional analyses reveal that GIFT maintains objective consistency throughout the two-stage post-training process. By preserving the exploration landscape, GIFT facilitates accelerated convergence and superior final performance during the subsequent RL stage, enhancing the model's reasoning capabilities.

The contributions of this work are twofold, spanning theoretical and practical dimensions. Theoretically, by leveraging the well-established Gibbs optimal policy from KL-regularized RL, we provide a novel reinterpretation of the post-training pipeline. Specifically, we show that standard SFT corresponds to the degenerate zero-temperature limit of this surrogate, which sharpens the policy and erodes the exploration space for subsequent RL. Practically, we propose GIFT to realize a principled finite-temperature initialization. Extensive experiments demonstrate that GIFT outperforms robust SFT baselines and matches competitive unified post-training methods on reasoning and OOD benchmarks, while maintaining superior objective consistency throughout the two-stage post-training pipeline.

\section{Related Work}
\paragraph{Post-training for Reasoning: SFT-then-RL Paradigm.} Post-training for reasoning LLMs typically follows a two-stage paradigm: Supervised Fine-Tuning (SFT) and Reinforcement Learning with Verifiable Rewards (RLVR) ~\cite{deepseekai2025deepseekr1incentivizingreasoningcapability, team2025kimi,liu2025acereasonnemotron11advancingmath}. SFT acts as a "cold-start" phase, establishing a strong initial policy by training on high-quality reasoning chains. Subsequently, RLVR enhances problem-solving capabilities by allowing the model to explore the solution space beyond the SFT data, discovering novel and more robust reasoning paths. However, this transition is impeded by an intrinsic optimization conflict~\cite{ouyang2022traininglanguagemodelsfollow, touvron2023llama2openfoundation}. Specifically, SFT typically compels the model to fit deterministic one-hot labels, which aggressively suppresses the probability mass of alternative tokens. In contrast, RL necessitates a high-entropy policy distribution to facilitate exploration and the discovery of high-reward trajectories~\cite{chen2025retainingdoingroleonpolicy, zhang2024mixcetrainingautoregressivelanguage, kumar2022finetuningdistortpretrainedfeatures, niu2026nondecouplingsupervisedfinetuningreinforcement}. This mismatch often causes SFT to overfit specific patterns and induce a collapse of the probability manifold, thereby stifling the exploration capacity required for the subsequent RL stage~\cite{kang2025quagmiressftrlposttraininghigh, vattikonda2025trainllmwebagent, chen2025stepwiseadaptiveintegrationsupervised}.

\paragraph{Refining SFT and Bridging Post-Training Stages.}
To mitigate the distributional degradation inherent in standard SFT, numerous studies have refined its objective by modifying the training loss (e.g., dynamic re-weighting and clipping \cite{li2025loglikelihoodprobabilitybasedobjectives, wu2025generalizationsftreinforcementlearning, zhu2025anchoredsupervisedfinetuning, zhu2025proximalsupervisedfinetuning}) or isolating loss calculations to critical high-entropy reasoning tokens \cite{wang20258020rulehighentropyminority, jiang2025rethinkingentropyregularizationlarge, ruan2025enhancinglargelanguagemodel}. However, treating SFT as an isolated task often fails to address its overarching optimization mismatch with subsequent RL. To bridge this gap, alternative efforts seek to interleave the two stages \cite{ma2025learningreinforcementlearningcant, MIXCHORD} or unify their objectives into a single framework \cite{yan2025learningreasonoffpolicyguidance, ming2025onetokenrolloutguidingsupervised, liu2025uftunifyingsupervisedreinforcement, chen2025bridgingsupervisedlearningreinforcement}. Despite their theoretical potential, these unified approaches introduce substantial computational complexity and have yet to supersede the empirically robust SFT-then-RL pipeline. In contrast, GIFT operates within this standard paradigm, proposing a lightweight, single-pass and mathematically principled initialization method designed to reconcile optimization objectives across the entire post-training process.

\input{latex/sections/method}

\section{Experiments}
\label{sec:experiments}
\subsection{Experimental setup}
\noindent \textbf{Datasets.} We utilize \textbf{DeepMath-103k}~\cite{he2025deepmath}, a large-scale mathematical dataset specifically curated for high-difficulty reasoning tasks. It undergoes a rigorous decontamination process against multiple public benchmarks and provides verifiable ground-truth answers to facilitate rule-based rewards. Each problem is accompanied by high-quality solutions generated by DeepSeek-R1 \cite{deepseekai2025deepseekr1incentivizingreasoningcapability}. From the full DeepMath-103k corpus, we randomly sample three disjoint subsets for our experiments: 10,000 samples for SFT, 10,000 samples for RL, and 1,000 samples as a validation set to monitor convergence during the SFT stage.


\noindent \textbf{Evaluation.} To comprehensively assess the reasoning capabilities, we employ a suite of 6 widely adopted benchmarks: mathematical reasoning benchmarks \textbf{GSM8K}~\cite{cobbe2021trainingverifierssolvemath} for grade-school math; \textbf{Math500}~\cite{hendrycks2021measuringmathematicalproblemsolving}, a rigorous subset of the MATH dataset; \textbf{AIME24} \& \textbf{AIME25} to test performance on high-difficulty mathematics competitions; and \textbf{HumanEval-plus}, \textbf{MBPP-plus} ~\cite{evalplus} for code reasoning. Furthermore, to verify generalizability, we evaluate on 4 OOD benchmarks: \textbf{GPQA}~\cite{rein2023gpqagraduatelevelgoogleproofqa} for graduate-level reasoning, \textbf{MMLU-Pro}~\cite{wang2024mmluprorobustchallengingmultitask} and \textbf{MMLU-Redux}~\cite{gema2025mmlu} for massive multitask language understanding,  \textbf{ARC-Challenge}~\cite{clark2018thinksolvedquestionanswering} for common-sense reasoning. We also validate GIFT on open-ended generation tasks (\textbf{MT-Bench}~\cite{Bai_2024}), see Appendix \ref{app:open-ended}.  All evaluations are conducted by VeRL~\cite{sheng2024hybridflow} with the vLLM backend. We set the maximum response length to 8,192 tokens and temperature $T=0.6$ to evaluate the robustness of the policy. We report pass@1 performance for all benchmarks.

\noindent \textbf{Baselines and Implementation.} We evaluate various initialization strategies on two backbone models, Qwen2.5-7B and Llama-3.1-8B, categorizing baselines into three groups: (1) \textbf{Direct SFT/RL}, which includes standard SFT on the base model and applying RL directly without prior fine-tuning; (2) \textbf{Unified Paradigms}, featuring recent methods like LUFFY ~\cite{yan2025learningreasonoffpolicyguidance} and ReLIFT ~\cite{ma2025learningreinforcementlearningcant}; and (3) \textbf{SFT-then-RL}, where we compare GIFT against Standard SFT and explicit ablations including SFT+Entropy (entropy regularization). To rigorously isolate the effect of our reward-calibrated Gibbs re-weighting, we also evaluate structurally similar target-softening approaches—SFT+Label Smoothing and SFT+KD—with their hyperparameters tuned to optimal capacity (see Appendix~\ref{app:regularization_sweep}). We also compare with recent SFT methods including DFT~\cite{wu2025generalizationsftreinforcementlearning}, ASFT~\cite{zhu2025anchoredsupervisedfinetuning}, and PSFT~\cite{zhu2025proximalsupervisedfinetuning}. Notably, we omit the results of Unified Paradigms on Llama-3.1-8B due to their consistently poor performance on this weaker backbone. In the subsequent RL stage, we uniformly employ GRPO~\cite{shao2024deepseekmathpushinglimitsmathematical} via the VeRL framework. Training is configured with a group size of $G=8$, a learning rate of $1\times 10^{-6}$, a PPO clip ratio of 0.2, and a maximum length of 8,192 tokens for long chain-of-thought reasoning. SFT is trained for 6 epochs and the subsequent RL stage runs for 1 epoch on 8$\times$NVIDIA H200 GPUs. Detailed hyperparameters and settings are provided in Appendix \ref{app:details}.

\subsection{Main Results}
\label{sec:main_results}
We demonstrate the effectiveness of GIFT initialization from three aspects: significant improvements in reasoning tasks, robust OOD generalization, and superior exploration potential as evidenced by pass@k scaling prior to the RL training.

\begin{table*}[t]
\scriptsize
\centering
\caption{Average pass@1 performance of different initialization strategies for subsequent RL on six reasoning benchmarks. The best result in each column is bold.}
\label{tab:math_results}
\resizebox{\textwidth}{!}{
\begin{tabular}{llccccccc}
\toprule
\textbf{Method Class} & \textbf{Strategy} & \textbf{GSM8K} & \textbf{AIME24} & \textbf{AIME25} & \textbf{MATH500} & \textbf{HE}$^{+}$ & \textbf{MBPP}$^{+}$ & \textbf{Average} \\
\midrule
\multicolumn{9}{c}{\textit{\textbf{Qwen2.5-7B}}} \\
\midrule
\multirow{2}{*}{\textit{Direct SFT/RL}} 
 & Direct SFT & 90.25 & 15.00 & 16.66 & 74.30 & 73.20 & 61.90 & 55.22 \\
 & Direct RL & 90.50 & 12.22 & 6.67 & 77.13 & 73.20 & 63.80 & 53.92 \\
\cmidrule{1-9}
\multirow{2}{*}{\textit{Unified Paradigms}} 
 & LUFFY & \textbf{92.70} & 15.00 & 15.83 & 80.20 & 73.20 & 63.20 & 56.69\\
 & ReLIFT & 91.84 & 14.06 & 17.50 & 79.72 & 72.60 & 61.90 & 56.27\\
\cmidrule{1-9}
\multirow{8}{*}{\textit{SFT-then-RL}} 
 & SFT & 91.79 & 13.33 & 14.44 & 78.67 & 72.00 & 60.10 & 55.06 \\
 & SFT + Entropy & 90.90 & 18.33 & 11.67 & 79.90 & 73.80 & 61.90 & 56.08 \\
 & SFT + Label Smoothing & 90.14 & 13.33 & 10.00 & 73.60 & 72.00 & 59.00 & 53.01 \\
 & SFT + KD & 91.85 & 13.33 & 15.56 & 78.83 & 72.60 & 60.50 & 55.45 \\
 & DFT & 86.50 & 6.67 & 8.89 & 65.87 & 73.20 & 64.30 & 50.91 \\
 & ASFT & 90.65 & 12.22 & 7.78 & 74.07 & 73.80 & 63.00 & 53.59 \\
 & PSFT & 91.94 & 15.83 & 16.67 & 81.33 & 73.20 & 59.00 & 56.33 \\
 & \textbf{GIFT} & 92.06 & \textbf{23.33} & \textbf{17.78} & \textbf{82.00} & \textbf{76.20} & \textbf{65.90} & \textbf{59.55} \\
\midrule
\multicolumn{9}{c}{\textit{\textbf{Llama3.1-8B}}} \\
\midrule
\multirow{2}{*}{\textit{Direct SFT/RL}} 
 & Direct SFT & 75.59 & \textbf{6.67} & 0.00 & 39.50 & 39.00 & 43.10 & 33.98 \\
 & Direct RL & 49.18 & 0.00 & 3.33 & 22.27 & 34.10 & 48.70 & 26.26 \\
\cmidrule{1-9}
\multirow{8}{*}{\textit{SFT-then-RL}} 
 & SFT & 73.39 & 0.00 & 3.33 & 39.20 & 43.30 & 45.50 & 34.12 \\
 & SFT + Entropy & 70.24 & \textbf{6.67} & 3.33 & 36.30 & 40.90 & 48.40 & 34.31 \\
 & SFT + Label Smoothing & 67.90 & 0.00 & 0.00 & 33.40 & 36.60 & 47.10 & 30.83 \\
 & SFT + KD & 73.48 & 0.00 & 3.33 & 39.52 & 43.00 & 45.80 & 34.19 \\
 & DFT & 64.75 & 0.00 & 3.33 & 29.47 & 40.90 & 49.50 & 31.33 \\
 & ASFT & 60.78 & 3.33 & 0.00 & 24.33 & 39.60 & 47.10 & 29.19 \\
 & PSFT & 74.53 & 3.33 & \textbf{6.67} & 39.40 & \textbf{44.50} & 44.40 & 35.47 \\
 & \textbf{GIFT} & \textbf{81.33} & \textbf{6.67} & \textbf{6.67} & \textbf{44.45} & \textbf{44.50} & \textbf{50.80} & \textbf{39.07} \\
\bottomrule
\end{tabular}
}
\end{table*}

\begin{table*}[t]

\centering

\scriptsize

\caption{Average pass@1 performance of different initialization strategies for subsequent RL on four out-of-distribution benchmarks. The best result in each column is bold.}

\label{tab:combined_ood}

\resizebox{\textwidth}{!}{

\begin{tabular}{llccccc}

\toprule

\textbf{Method Class} & \textbf{Strategy} & \textbf{GPQA} & \textbf{MMLU-Redux} & \textbf{ARC-Challenge} & \textbf{MMLU-Pro} & \textbf{Average} \\

\midrule

\multicolumn{7}{c}{\textit{\textbf{Qwen2.5-7B}}} \\

\midrule

\multirow{2}{*}{\textit{Direct SFT/RL}} 

 & Direct SFT & 19.70 & 69.03 & 82.68 & 50.23 & 55.41 \\

 & Direct RL & 32.32 & 70.50 & 73.85 & 54.33 & 57.75 \\

\cmidrule{1-7}


\multirow{2}{*}{\textit{Unified Paradigms}} 

 & LUFFY & 34.85 & 73.30 & 89.85 & \textbf{59.08} & 64.27 \\

 & ReLIFT & 37.88 & 73.67 & \textbf{89.85} & 56.82 & \textbf{64.55} \\

\cmidrule{1-7}

\multirow{6}{*}{\textit{SFT-then-RL}} 

 & SFT & 27.78 & 71.13 & 85.54 & 54.67 & 59.78 \\

 & SFT + Entropy & 33.33 & 72.20 & 87.29 & 55.60 & 62.11 \\

 & DFT & 17.17 & 57.77 & 84.85 & 39.75 & 49.89 \\

 & ASFT & 25.25 & 68.27 & 86.90 & 49.98 & 57.60 \\

 & PSFT & 34.34 & \textbf{73.79} & 87.77 & 59.05 & 63.74 \\

 & \textbf{GIFT} & \textbf{39.09} & 73.27 & 88.01 & 56.01 & 64.10 \\

\midrule

\multicolumn{7}{c}{\textit{\textbf{Llama3.1-8B}}} \\

\midrule

\multirow{2}{*}{\textit{Direct SFT/RL}} 

 & Direct SFT & 21.21 & 61.50 & 77.82 & 40.05 & 50.14 \\

 & Direct RL & 23.74 & 52.13 & 64.59 & 29.65 & 42.53 \\

\cmidrule{1-7}

\multirow{6}{*}{\textit{SFT-then-RL}} 

 & SFT & 23.74 & 55.97 & 74.49 & 33.67 & 46.97 \\

 & SFT + Entropy & 27.78 & \textbf{63.37} & 80.25 & 41.87 & 53.32 \\

 & DFT & 24.24 & 54.27 & 71.46 & 35.59 & 46.39 \\

 & ASFT & 13.13 & 47.23 & 66.42 & 29.40 & 39.05 \\

 & PSFT & 12.12 & 44.67 & 46.54 & 30.14 & 33.37 \\

 & \textbf{GIFT} & \textbf{30.81} & 63.17 & \textbf{82.94} & \textbf{44.05} & \textbf{55.24} \\

\bottomrule

\end{tabular}

}

\end{table*}

\noindent \textbf{Reasoning Performance.} As shown in Table \ref{tab:math_results}, GIFT consistently achieves the highest average performance across all initialization strategies on both Qwen2.5-7B and Llama-3.1-8B. On Qwen2.5-7B, GIFT attains an average pass@1 of 59.55\% (+4.49\% over SFT), surpassing not only strong SFT variants like PSFT (56.33\%) but also complex unified paradigms such as LUFFY (56.69\%). Notably, on the challenging AIME benchmark, GIFT yields a substantial gain of 10\% over Standard SFT (13.33\% $\rightarrow$ 23.33\%). On Llama-3.1-8B, GIFT similarly leads with an average of 39.07\%, outperforming Standard SFT (34.12\%) and PSFT (35.47\%) by clear margins. In contrast, while simple regularization via SFT + Entropy yields moderate improvements, baselines such as DFT and ASFT exhibit severe degradation. Notably, even with the hyperparameter sweep in Appendix~\ref{app:regularization_sweep}, the best-tuned KD trails GIFT by 4.10\%/4.88\% on Qwen2.5-7B/Llama-3.1-8B, while Label Smoothing strictly degrades performance. This isolates the effect of GIFT's reward-calibrated anchor, which explicitly boosts the oracle token while preserving the base model's relative probability mass.

\noindent \textbf{Generalization Capabilities.} Table \ref{tab:combined_ood} details the OOD performance across different initialization strategies. GIFT demonstrates robust generalization capabilities, effectively mitigating the catastrophic forgetting of general knowledge typically induced by the rigid zero-temperature collapse of Standard SFT and other variants. On Qwen2.5-7B, GIFT achieves an average score of 64.10\%, clearly outperforming Standard SFT (59.78\%) and on par with complex Unified Paradigms like ReLIFT (64.55\%). On Llama-3.1-8B, GIFT secures the highest average performance (55.24\%), surpassing all baselines. These results confirm that by maintaining a finite temperature, GIFT successfully retains the structural priors essential for broad generalization, whereas standard zero-temperature SFT is prone to overfitting.

\noindent \textbf{Exploration Potential.} Pre-RL pass@$k$ evaluation (Table \ref{tab:pass_at_k}) shows that GIFT scales more favorably with the sample budget. On Qwen2.5-7B, standard SFT holds a minor pass@1 advantage, but GIFT extends the lead to +5.16\% by pass@8. On Llama-3.1-8B—which exhibits more rapid SFT overfitting—GIFT closes its initial deficit and performs on par with SFT at pass@8. While Llama's pre-RL pass@$k$ saturates early, the broader output distribution retained by GIFT becomes useful during subsequent RL (Table \ref{tab:math_results}), where it supports the downstream gains. Detailed breakdowns are in Appendix \ref{sec:detailed_results}.

\begin{table}[h]
\centering
\small
\caption{Average pass@$k$ performance (\%) on four mathematical reasoning benchmarks in SFT stage prior to RL.}
\label{tab:pass_at_k}
\resizebox{\linewidth}{!}{
\begin{tabular}{llcccc}
\toprule
\textbf{Model} & \textbf{Strategy} & \textbf{$k=1$} & \textbf{$k=2$} & \textbf{$k=4$} & \textbf{$k=8$} \\
\midrule
\multirow{2}{*}{Llama-3.1-8B}
 & SFT & \textbf{27.83} & 34.35 & \textbf{39.67} & 44.25 \\
 & \textbf{GIFT} & 27.60 & \textbf{34.53} & 39.62 & \textbf{44.87} \\
\midrule
\multirow{2}{*}{Qwen2.5-7B}
 & SFT & \textbf{46.64} & 52.16 & 55.90 & 58.02 \\
 & \textbf{GIFT} & 46.06 & \textbf{53.92} & \textbf{58.48} & \textbf{63.18} \\
\bottomrule
\end{tabular}
}
\end{table}

\subsection{Geometric and Distributional Consistency}
\label{sec:geometric_analysis}
To validate whether the finite-temperature initialization fosters a more continuous and stable optimization trajectory during post-training, we analyze the geometric and distributional properties of the model updates. We evaluate the model's representational and distributional shifts across two sequential stages: \textbf{(1)} the transition from the base model to the initialized policy (\textit{Base $\to$ SFT}), and \textbf{(2)} the subsequent evolution from the initialized policy to the final RL model (\textit{SFT $\to$ RL}). We quantify geometric consistency using cosine similarity and L2 distance of the last transformer layer, while measuring distributional consistency via KL divergence and Top-$K$ token overlap (estimated on the held-out training subset). While smaller representational shifts are not unconditionally beneficial in post-training, they serve as useful proxy metrics in our context to indicate the preservation of structural priors.

\begin{table}[t]
\centering
\caption{Consistency Metrics across Training Stages.}
\label{tab:main_consistency}
\resizebox{\linewidth}{!}{%
\begin{tabular}{l|ccc|ccc}
\toprule
 & \multicolumn{3}{c|}{\textbf{Stage 1 Base $\to$ SFT}} & \multicolumn{3}{c}{\textbf{Stage 2 SFT $\to$ RL}} \\
\textbf{Strategy} & \textbf{Cos} $\uparrow$ & \textbf{L2} $\downarrow$ & \textbf{KL} $\downarrow$ & \textbf{Cos} $\uparrow$ & \textbf{L2} $\downarrow$ & \textbf{KL} $\downarrow$ \\
\midrule
\multicolumn{7}{c}{\textit{\textbf{Llama-3.1-8B}}} \\
SFT & 0.8112 & 87.33 & \textbf{0.5893} & 0.9310 & 53.85 & 0.2316 \\
\textbf{GIFT} & \textbf{0.8207} & \textbf{85.05} & 0.5909 & \textbf{0.9422} & \textbf{49.08} & \textbf{0.2010} \\
\midrule
\multicolumn{7}{c}{\textit{\textbf{Qwen2.5-7B}}} \\
SFT & 0.9201 & 117.18 & 0.3995 & 0.9764 & 62.61 & 0.0426 \\
\textbf{GIFT} & \textbf{0.9253} & \textbf{113.95} & \textbf{0.3907} & \textbf{0.9869} & \textbf{47.34} & \textbf{0.0327} \\
\bottomrule
\end{tabular}
}
\end{table}

\noindent \textbf{Geometric Consistency.}
As reported in Table \ref{tab:main_consistency}, GIFT significantly enhances geometric consistency throughout the post-training process. GIFT consistently exhibits higher cosine similarity and lower L2 distance compared to standard SFT across stages and models. This indicates that GIFT effectively absorbs task-specific supervision while adhering to the base model's semantic manifold, reducing the destructive parameter updates caused by standard SFT, thereby providing a more efficient exploration for the subsequent RL stage.

\begin{figure*}[htbp] 
    \centering
    \includegraphics[width=0.9\textwidth]{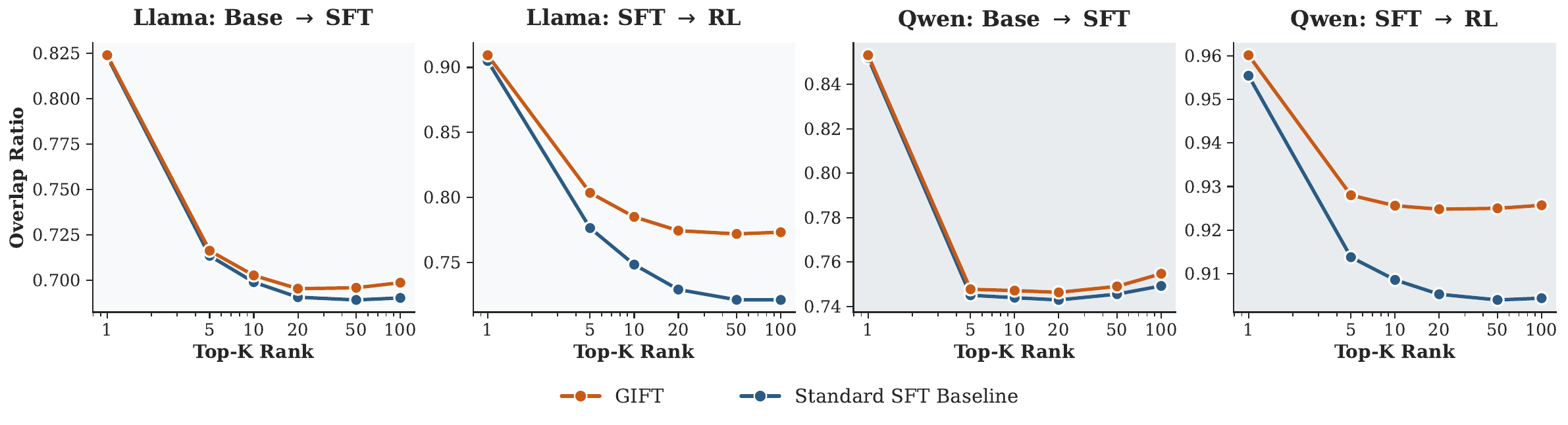}
    \caption{\textbf{Top-$K$ Token Overlap Analysis.} 
    Results for Llama-3.1-8B (Left) and Qwen2.5-7B (Right). GIFT exhibits consistently higher token overlap than the SFT baseline.}
    \label{fig:overlap}
\end{figure*}

\noindent \textbf{Distributional Consistency.}
As shown in Table \ref{tab:main_consistency}, GIFT reduces KL divergence across the entire post-training pipeline in most cases. Complementing this, the top-$k$ token overlap analysis in Figure \ref{fig:overlap} reveals that GIFT mitigates the drastic distributional shifts characteristic of standard SFT. Moreover, we can observe that the preservation of base priors during SFT stage is effectively leveraged in the subsequent RL stage, manifesting as higher top-$k$ overlap that facilitates more efficient exploration. These proxy metrics suggest that GIFT mitigates drastic distributional shifts, maintaining optimization coherence without forcing a rigid collapse.

\subsection{Analysis of Inverse Temperature}
\label{sec:beta_analysis}

\begin{figure*}[t]
    \centering
    \includegraphics[width=0.83\textwidth]{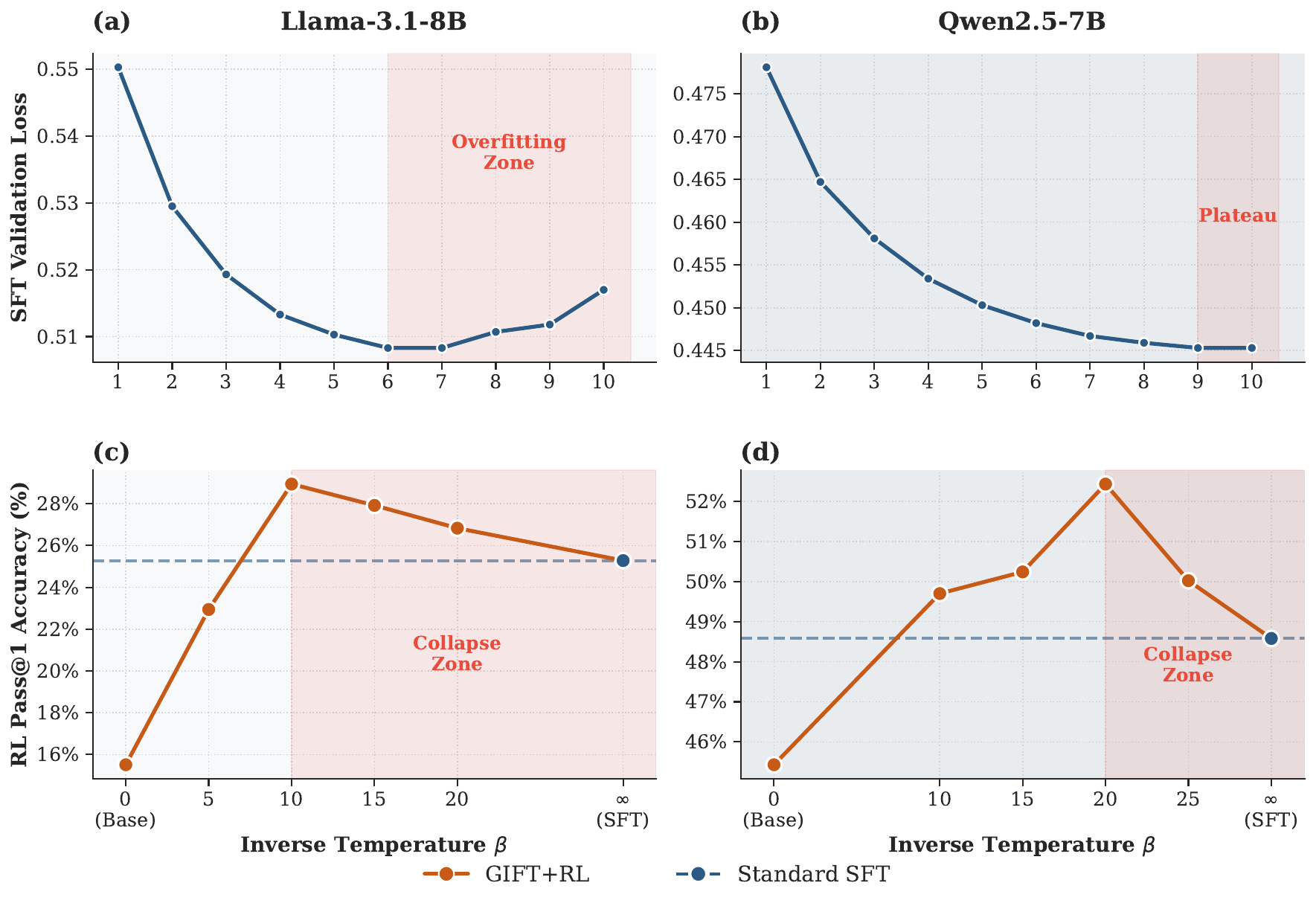}
    \caption{\textbf{Correlation between SFT validation dynamics and optimal RL initialization.} \textbf{(a)} and \textbf{(b)} show the SFT validation loss per epoch for Llama-3.1-8B and Qwen2.5-7B, respectively. \textbf{(c)} and \textbf{(d)} illustrate their corresponding post-RL pass@1 accuracy on mathematical reasoning benchmarks across different inverse temperatures $\beta$.}
    \label{fig:beta}
\end{figure*}

\begin{table}[t]
\centering
\scriptsize
\caption{Post-RL residual drift $\widehat{\eta}_{\text{final}}$ evaluated on a held-out training subset. All policies underwent the same RL phase but differ in their stage-one initialization (Standard SFT vs.\ GIFT at varying $\beta$).}
\label{tab:eta_final}
\resizebox{\linewidth}{!}{%
\begin{tabular}{lcccccc}
\toprule
\textbf{Model} & \textbf{SFT} & $\beta{=}5$ & $\beta{=}10$ & $\beta{=}15$ & $\beta{=}20$ & $\beta{=}25$ \\
\midrule
Qwen2.5-7B   & 0.030 & --- & 0.014 & 0.013 & \textbf{0.009} & 0.019 \\
Llama-3.1-8B & 0.044 & 0.059 & \textbf{0.016} & 0.047 & 0.040 & --- \\
\bottomrule
\end{tabular}}
\end{table}

The inverse temperature $\beta$ regulates the trade-off between expert imitation and prior preservation. As shown in Figure~\ref{fig:beta}, post-RL performance consistently peaks at a finite $\beta$. This optimal value correlates closely with a backbone's resilience to SFT overfitting: Llama-3.1-8B overfits early (Figure~\ref{fig:beta}a) and peaks at a lower $\beta \approx 10$, whereas Qwen2.5-7B maintains a stable validation loss (Figure~\ref{fig:beta}b) and tolerates a higher $\beta \approx 20$ before degrading toward the standard SFT baseline ($\beta \to \infty$).

To understand how the alignment effort is distributed across the two stages, we estimate the total parameter drift from the base model using $\widehat{\eta}_{\text{final}}(\beta)$, defined as the within-prompt difference in length-normalized log-likelihood ratios—relative to the base prior—between successful and failed rollouts on a held-out training subset. As shown in Table~\ref{tab:eta_final}, this cumulative drift exhibits a distinct U-shape, reaching its minimum near the same $\beta$ that maximizes downstream accuracy. This reflects the joint optimization dynamics of the pipeline. When $\beta$ is excessively large, the SFT stage aggressively overfits the demonstrations, introducing a large initial drift into the model. The RL stage inherits this collapsed state, keeping the final cumulative drift artificially inflated. Conversely, when $\beta$ is too small, the SFT stage offers insufficient guidance, forcing the RL stage to absorb the alignment burden and independently drive up the drift. The minimum of $\widehat{\eta}_{\text{final}}$ thus marks a state of maximum efficiency, where the SFT target better conditions the policy without breaking the underlying prior.

Consequently, the ideal $\beta$ admits a two-step localization: the onset of SFT validation overfitting (Figure~\ref{fig:beta}a,b) provides a pre-RL upper bound that already excludes the high-$\beta$ regime, and within this range $\widehat{\eta}_{\text{final}}$ further pinpoints the operating point where alignment effort is most evenly absorbed across the two stages. In our main experiments, we fix $\beta=20$ for Qwen2.5-7B and $\beta=10$ for Llama-3.1-8B based on a small pilot sweep on each backbone (Figure~\ref{fig:beta}c,d), and keep these values identical across all reported benchmarks.

\section{Conclusion}
In this work, we identify the zero-temperature distributional collapse induced by standard SFT and propose GIFT as a principled initialization for subsequent RL. Experiments validate its effectiveness across reasoning and out-of-distribution tasks. By reconciling the traditional two-stage paradigm, our work offers a theoretically motivated pathway toward constructing robust and generalizable reasoning systems.

\section{Limitations}
While our framework demonstrates strong performance on models such as Llama-3.1-8B and Qwen2.5-7B, evaluating its scaling properties on massive-scale LLMs is currently constrained by substantial computational requirements. Our current formulation is particularly well suited to sparse, verifiable-reward post-training, where demonstrations provide high-quality trajectories and the subsequent RL stage benefits from retaining base-model alternatives. Furthermore, our empirical validation primarily focuses on standard benchmarks within specific linguistic boundaries. Extending this policy optimization approach to highly specialized domains or comprehensive multilingual settings remains a natural avenue for future work.

\section*{Acknowledgments}
This work is supported by Fundamental and Interdisciplinary Disciplines Breakthrough Plan of the Ministry of Education of China (JYB2025XDXM113), National Natural Science Foundation of China (92470121, 62402016), National Key R\&D Program of China (2024YFA1014003), Zhongguancun Academy (C20250204, C20250602), Beijing Major Science and Technology Project (Z251100008125043, Z251100008425023), and High-performance Computing Platform of Peking University.

\bibliography{acl_latex} 

\clearpage
\input{latex/appendix}

\end{document}

%% file: latex/sections/method.tex
\section{Method}
\label{sec:method}
We propose \textbf{G}ibbs \textbf{I}nitialization with \textbf{F}inite \textbf{T}emperature (\textbf{GIFT}), an SFT target designed to mathematically bridge the gap between initial fine-tuning and subsequent RL. Rather than treating SFT in isolation, we derive GIFT by additively decomposing the overarching post-training objective across the two stages (Section~\ref{sec:two_stage_bridge}). Concretely, GIFT is the closed-form optimizer of a per-token variational surrogate that up-weights the demonstration token while preserving the base distribution as a prior (Section~\ref{sec:token_var}). Parameterized by an inverse temperature $\beta$, it recovers standard one-hot SFT in the $\beta \to \infty$ degenerate limit. The training algorithm is summarized in Section~\ref{sec:algorithm}.

\subsection{Preliminaries}
\noindent \textbf{Supervised Fine-Tuning.} Starting with a pre-trained base model $\pi_{\text{base}}$ and a curated dataset of high-quality demonstrations $\mathcal{D} = \{(x, y^*)\}$, SFT optimizes the model parameters $\theta$ to maximize the log-likelihood of the target sequences. The objective is typically formulated as a cross-entropy loss:
\begin{equation}
    \begin{split}
        \mathcal{L}_{\text{SFT}}(\theta) &= - \mathbb{E}_{(x, y^*) \sim \mathcal{D}} \log \pi_\theta(y^* | x) \\
        &= - \mathbb{E}_{(x, y^*) \sim \mathcal{D}} \sum_{t=1}^{|y^*|} \left[\log \pi_\theta(y^*_{t} | x,y_{<t}^*) \right].
    \end{split}
\end{equation}
By minimizing this negative log-likelihood, the model learns to imitate the distribution of the expert data, providing a specialized foundation for subsequent alignment or reasoning-based stages.

\noindent \textbf{Reinforcement Learning with Verifiable Rewards (RLVR).} Following SFT, the model $\pi_{\text{sft}}$ is further refined using RL algorithms—such as PPO \cite{schulman2017proximal} or GRPO \cite{shao2024deepseekmath}—to maximize a KL-regularized reward objective:
\begin{equation}
\label{eq:RL_obj}
\begin{split}
    J_{\text{RL}}(\theta) &= \mathbb{E}_{x \sim \mathcal{D},\, y \sim \pi_\theta} \Big[ R(x, y) \\
    &\quad - \lambda \, D_{\text{KL}}\big(\pi_\theta(\cdot|x) \,\big\|\, \pi_{\text{sft}}(\cdot|x)\big) \Big],
\end{split}
\end{equation}
where $\lambda > 0$ is the KL regularization coefficient. The KL anchor at $\pi_{\text{sft}}$ keeps the optimized policy close to the SFT initialization. This stage is critical for incentivizing advanced reasoning capabilities in LLMs, aiming to achieve superior performance that transcends the inherent limitations of static SFT datasets.

\subsection{A Variational Bridge Between SFT and RL}
\label{sec:two_stage_bridge}
A critical limitation of current post-training pipelines~\cite{vattikonda2025trainllmwebagent,kang2025quagmiressftrlposttraininghigh,ma2025learningreinforcementlearningcant} is the structural mismatch between the SFT outcome and the RL prerequisite. Standard SFT optimizes maximum likelihood, which inherently drives the policy toward a highly peaked, degenerate distribution around the expert demonstrations. However, the subsequent RL phase relies on a KL divergence penalty $D_{\text{KL}}(\pi_\theta \,\|\, \pi_{\text{sft}})$ to regularize exploration. If the anchor $\pi_{\text{sft}}$ is overly collapsed, this KL penalty heavily penalizes any exploratory deviation, severely limiting the RL optimization. 

Consequently, the SFT objective must not be treated as an isolated imitation task. Instead, it should be instantiated as an initialization that is structurally aligned with the KL-regularized form of the subsequent RL stage. We start by analyzing the target destination. In KL-regularized RL (Eq.~\ref{eq:RL_obj}), the inverse of the KL coefficient ($1/\lambda$) acts as an inverse temperature, dictating the strength of reward maximization. Viewing the entire pipeline as a unified process to reach an ultimate equilibrium, we govern its overarching goal with an \textit{aggregate inverse temperature} $\bar{\eta}$. Following the autoregressive factorization with base prior $\pi_{\text{base},t}$ and expert indicator $r_t(y) := \mathbb{I}[y = y^*_t]$, we seek a conditional distribution $q_t \in \Delta(\mathcal{V})$ that balances the expected reward against a KL penalty. This balance is formalized by the aggregate per-token variational free energy:
\begin{equation}
\label{eq:aggregate_free_energy}
\mathcal{F}_{\bar{\eta}}(q_t) = \bar{\eta} \, \mathbb{E}_{y \sim q_t}\!\left[ r_t(y) \right] - D_{\text{KL}}\!\big(q_t \,\big\|\, \pi_{\text{base},t}\big).
\end{equation}

We establish the mathematical bridge across this divide by decomposing the overarching objective. For any intermediate inverse temperature $0 \leq \beta \leq \bar{\eta}$, it yields an exact additive identity:
\begin{equation}
\label{eq:decomposition}
\mathcal{F}_{\bar{\eta}}(q_t) = \mathcal{F}_\beta(\pi_{\text{sft}}^*) + \mathcal{F}^{(2)}_{\bar{\eta}-\beta}\!\big(q_t;\, \pi_{\text{sft}}^*\big),
\end{equation}
where $\pi_{\text{sft}}^*$ is the optimal target at $\beta$, and the residual free energy is:
\begin{equation}
\label{eq:residual}
\mathcal{F}^{(2)}_{\bar{\eta}-\beta}\!\big(q_t;\, \pi_{\text{sft}}^*\big) := (\bar{\eta} - \beta) \, \mathbb{E}_{q_t}[r_t] - D_{\text{KL}}\!\big(q_t \,\|\, \pi_{\text{sft}}^*\big).
\end{equation}

This decomposition resolves the structural disconnect. The residual term structurally mirrors the RL objective anchored at $\pi_{\text{sft}}^*$, operating under the remaining inverse temperature $(\bar{\eta}-\beta)$. Consequently, the first term $\mathcal{F}_\beta(\pi_{\text{sft}}^*)$ dictates that the SFT stage should not blindly maximize likelihood, but serve as a finite-temperature target that absorbs a portion ($\beta$) of the alignment pressure without triggering the prior collapse that limits RL. The exact algebraic validity of this two-stage additive decomposition is formally proven in Appendix~\ref{app:two_stage_proof}.

\subsection{Deriving the SFT Target}
\label{sec:token_var}
Guided by the decomposition above, the SFT stage must act to maximize the stage-one objective $\mathcal{F}_\beta(q_t) = \beta \mathbb{E}_{q_t}[r_t] - D_{\text{KL}}(q_t \| \pi_{\text{base},t})$. The variational problem $\max_{q_t \in \Delta(\mathcal{V})} \mathcal{F}_\beta(q_t)$ admits a closed-form solution. By Donsker--Varadhan duality~\cite{donsker1975asymptotic}, the supremum equals the log partition function:
\begin{equation}
\label{eq:dv}
\begin{aligned}
\sup_{q_t \in \Delta(\mathcal{V})} \mathcal{F}_\beta(q_t) &= \log Z_t(\beta), \\
\text{where} \quad Z_t(\beta) &:= \mathbb{E}_{\pi_{\text{base},t}}\!\big[\mathrm{e}^{\beta r_t}\big],
\end{aligned}
\end{equation}
and is attained at the closed-form maximizer:
\begin{equation}
\label{eq:our_method}
\pi_{\text{sft}}^*(y_t \mid x, y^*_{<t}) = \frac{1}{Z_t(\beta)} \, \pi_{\text{base},t}(y_t) \, \mathrm{e}^{\beta\, r_t(y_t)},
\end{equation}
which we adopt as the formal SFT target. Substituting the indicator form of $r_t$, Eq.~\ref{eq:our_method} reduces to the explicit two-case expression:
\begin{equation}
\label{eq:our_method_two_case}
\pi_{\text{sft}}^*(y \mid x, y^*_{<t}) =
\begin{cases}
\dfrac{\mathrm{e}^{\beta}\, \pi_{\text{base},t}(y^*_t)}{Z_t(\beta)} & y = y^*_t, \\[4pt]
\dfrac{\pi_{\text{base},t}(y)}{Z_t(\beta)} & y \neq y^*_t,
\end{cases}
\end{equation}
with normalizer $Z_t(\beta) = (\mathrm{e}^{\beta} - 1)\, \pi_{\text{base},t}(y^*_t) + 1$. On the logits level, this simply amounts to adding a constant bias $\beta$ at the demonstration's choice and re-normalizing. 

This formulation fundamentally clarifies the failure mode of standard SFT. Taking the limit $\beta \to \infty$ sends the mass at $y^*_t$ to one:
\begin{equation}
\label{eq:our_method_collapse}
\lim_{\beta \to \infty} \pi_{\text{sft}}^*(\,\cdot\, \mid x, y^*_{<t}) \;=\; \delta_{y^*_t},
\end{equation}
recovering standard SFT as a degenerate zero-temperature limit. Crucially, as $\pi_{\text{sft}}^* \to \delta_{y^*_t}$, the residual penalty $D_{\text{KL}}(q_t \| \pi_{\text{sft}}^*)$ in Eq.~\ref{eq:residual} diverges for any $q_t$ that places mass off $y^*_t$. The residual feasible region contracts strictly to $\{\delta_{y^*_t}\}$, restricting the search space available for the subsequent RL stage to refine the policy. Finite $\beta$ prevents this premature policy collapse, maintaining support across the vocabulary and ensuring a smooth allocation of the aggregate inverse temperature. In practice, we treat $\beta$ as a hyperparameter rather than tying it strictly to $\bar{\eta}$, accounting for trust-region clipping in modern RL (examined empirically in Section~\ref{sec:beta_analysis}).

\subsection{Proposed Algorithm: GIFT}
\label{sec:algorithm}
Minimizing the KL divergence to our variational target $\pi_{\text{sft}}^*$ (Eq.~\ref{eq:our_method}) is mathematically equivalent to optimizing the cross-entropy loss:
\begin{equation}   
    \begin{split}
        \mathcal{L}(\theta) = & - \mathbb{E}_{(x,y^*)\sim \mathcal{D}} \Big[\sum_{t=1}^{|y^*|}\sum_{y_t\in\mathcal{V}} \\
        & \pi_{\text{sft}}^*(y_t \mid x,y^*_{<t}) \log \pi_\theta(y_t \mid x,y^*_{<t})\Big].
    \end{split}
\end{equation}
The complete training procedure is outlined in Algorithm~\ref{alg:GIFT}. Unlike standard soft-targets, GIFT employs an exponential tilt, $q(y) \propto \pi_{\text{base}}(y)\exp(\beta \mathbb{I}[y=y^*])$, which preserves the relative odds among non-oracle tokens (detailed in Appendix~\ref{app:regularization_sweep}).

\begin{algorithm}[t]
\caption{GIFT: Gibbs Initialization with Finite Temperature}
\label{alg:GIFT}
\begin{algorithmic}[1]
\REQUIRE Data $\mathcal{D}$, base model $\pi_{\text{base}}$, inverse temperature $\beta$, learning rate $\alpha$
\STATE Initialize $\pi_\theta \leftarrow \pi_{\text{base}}$
\WHILE{not converged}
    \STATE Sample $(x, y^*)\sim\mathcal{D}$; obtain base log-probs $\ell_{t,k} \!=\! \log\pi_{\text{base}}(k|x,y^*_{<t})$
    \STATE Form GIFT target logits \\ $\hat z_{t,k} \!=\! \ell_{t,k} + \beta\,\mathbb{I}[k\!=\!y^*_t]$,\\then $\pi^*_{\text{sft},t} \!=\! \mathrm{Softmax}(\hat z_t)$
    \STATE $\mathcal{L} \leftarrow -\sum_{t,k} \pi^*_{\text{sft},t}(k)\,\log \pi_\theta(k|x,y^*_{<t})$
    \STATE $\theta \leftarrow \theta - \alpha\, \nabla_\theta \mathcal{L}$
\ENDWHILE
\RETURN $\pi_\theta$
\end{algorithmic}
\end{algorithm}

%% file: latex/appendix.tex
\newpage
\appendix

    
    

\section{Closed-Form Reference for KL-Regularized RL}
\label{appendix:derivation}

We record here the standard variational closed form of the KL-regularized RL maximizer used as a reference throughout the main text. For an arbitrary anchor $\pi_{\text{ref}}$ and KL coefficient $\lambda > 0$, the objective $J(\pi) = \mathbb{E}\!\left[ R(x,y) - \lambda\, D_{\text{KL}}\!\big(\pi(\cdot|x)\,\big\|\,\pi_{\text{ref}}(\cdot|x)\big) \right]$ is uniquely maximized at the Gibbs distribution
\begin{equation}
\label{eq:app_seq_gibbs}
\pi^*(y|x) \;\propto\; \pi_{\text{ref}}(y|x)\, \mathrm{e}^{R(x,y)/\lambda},
\end{equation}
with inverse temperature $1/\lambda$ on the reward. We invoke this expression at two points in the main text: (i) Eq.~\ref{eq:RL_obj}, with $\pi_{\text{ref}}=\pi_{\text{sft}}$, where it characterizes the equilibrium of the standard RLVR stage; and (ii) Section~\ref{sec:two_stage_bridge}, where the residual term in Eq.~\ref{eq:residual} is read as a KL-regularized RL problem anchored at $\pi_{\text{sft}}^*$ with $\lambda = 1/(\bar{\eta} - \beta)$; the corresponding identification of $\bar{\eta} - \beta$ as the inverse temperature contributed by the RL stage is made formal in Appendix~\ref{app:two_stage_proof}. While the above characterizes the sequence-level RL equilibrium, the GIFT target itself is derived natively at the token level from a per-token variational free energy, as detailed next in Appendix~\ref{app:token_dv}.

\section{Derivation of the GIFT Token-Level Target}
\label{app:token_dv}

This section provides the detailed derivation of the closed-form maximizer of the per-token variational free energy used in Section~\ref{sec:token_var}. Throughout, we drop the conditioning on $(x, y^*_{<t})$ and write $\pi_{\text{base}}(\cdot)$ for $\pi_{\text{base},t}(\cdot)$, $r(y) := r_t(y) = \mathbb{I}[y = y^*_t]$ for the per-token oracle indicator carried by the SFT demonstration $(x, y^*)$, and $q$ for the candidate token-level distribution. The variational problem is
\begin{equation}
\label{eq:app_var_prob}
\max_{q \in \Delta(\mathcal{V})} \; \mathcal{F}_\beta(q) = \beta \, \mathbb{E}_q[r] - D_{\text{KL}}\!\big(q \,\big\|\, \pi_{\text{base}}\big),
\end{equation}
with $\Delta(\mathcal{V})$ the simplex over the vocabulary $\mathcal{V}$.

\paragraph{Remark on the indicator form of $r$.} Setting $r(y) = \mathbb{I}[y = y^*_t]$ is the formal encoding of the supervision actually carried by an SFT pair $(x, y^*)$: at position $t$, only the demonstration token is identified, with no token-level credit available for any alternative $y \neq y^*_t$. The derivation below applies verbatim to an arbitrary bounded $r:\mathcal{V}\to\mathbb{R}$, in which case $\pi_{\text{sft}}^*(y) \propto \pi_{\text{base}}(y) \mathrm{e}^{\beta r(y)}$ is the corresponding maximizer; the indicator specialization is a structural necessity strictly imposed by standard SFT datasets: because the data only provides the single demonstrated sequence, we inherently lack the explicit supervisory signal to evaluate any alternative tokens.

\paragraph{Donsker--Varadhan route.} The Donsker--Varadhan variational characterization of the log-partition function states that
\begin{equation}
\log \mathbb{E}_{\pi_{\text{base}}}\!\big[\mathrm{e}^{\beta r}\big] = \sup_{q \in \Delta(\mathcal{V})} \; \mathcal{F}_\beta(q),
\end{equation}
with the supremum attained at the Gibbs distribution
\begin{equation}
\label{eq:app_token_gibbs}
\begin{split}
\pi_{\text{sft}}^*(y) &= \frac{1}{Z_t(\beta)} \, \pi_{\text{base}}(y) \, \mathrm{e}^{\beta\, r(y)}, \\
Z_t(\beta) &= \mathbb{E}_{\pi_{\text{base}}}\!\big[\mathrm{e}^{\beta r}\big].
\end{split}
\end{equation}

\paragraph{Lagrangian route.} The same maximizer follows from a direct Lagrangian argument. Introducing a multiplier $\nu$ for the normalization constraint $\sum_y q(y) = 1$, the Lagrangian is
\begin{equation*}
\begin{split}
\mathcal{L}(q, \nu) &= \beta \sum_y q(y) r(y) - \sum_y q(y) \log \frac{q(y)}{\pi_{\text{base}}(y)} \\
&\quad - \nu \Big(\sum_y q(y) - 1\Big).
\end{split}
\end{equation*}
Differentiating with respect to $q(y)$ and setting the result to zero yields
\begin{equation*}
\beta \, r(y) - \log q(y) + \log \pi_{\text{base}}(y) - 1 - \nu = 0,
\end{equation*}
so $q(y) \propto \pi_{\text{base}}(y) \mathrm{e}^{\beta r(y)}$, which after normalization recovers Eq.~\ref{eq:app_token_gibbs}. Strict concavity of $-D_{\text{KL}}(\cdot \| \pi_{\text{base}})$ on the relative interior of $\Delta(\mathcal{V})$, together with the linearity of $\beta \mathbb{E}_q[r]$, ensures that the maximizer is unique whenever $\pi_{\text{base}}$ has full support on $\mathcal{V}$.

\paragraph{Specialization to the per-token oracle.} Substituting $r(y) = \mathbb{I}[y = y^*_t]$, the partition function simplifies to
\begin{equation*}
Z_t(\beta) = (\mathrm{e}^{\beta} - 1) \, \pi_{\text{base}}(y^*_t) + 1,
\end{equation*}
and $\pi_{\text{sft}}^*$ admits the explicit two-case form
\begin{equation*}
\pi_{\text{sft}}^*(y) = \begin{cases}
\dfrac{\mathrm{e}^{\beta}\, \pi_{\text{base}}(y^*_t)}{Z_t(\beta)} & \text{if } y = y^*_t, \\[6pt]
\dfrac{\pi_{\text{base}}(y)}{Z_t(\beta)} & \text{if } y \neq y^*_t,
\end{cases}
\end{equation*}
matching Eq.~\ref{eq:our_method_two_case} of the main text. At the level of logits, this is equivalent to adding a constant $\beta$ to the base logit at $y^*_t$ and re-applying softmax, which is exactly the target constructed in Algorithm~\ref{alg:GIFT}. The cross-entropy loss in Algorithm~\ref{alg:GIFT} therefore fits $\pi_\theta(\cdot \mid x, y^*_{<t})$ to the closed-form maximizer of Eq.~\ref{eq:app_var_prob}.

\paragraph{The $\beta \to \infty$ limit.} For finite $\beta$, $\pi_{\text{sft}}^*$ retains support on the entire vocabulary $\mathcal{V}$. As $\beta \to \infty$, $\pi_{\text{sft}}^*(y^*_t) = \mathrm{e}^{\beta}\pi_{\text{base}}(y^*_t)/Z_t(\beta) \to 1$ and $\pi_{\text{sft}}^*(y) \to 0$ pointwise for every $y \neq y^*_t$, so $\pi_{\text{sft}}^* \to \delta_{y^*_t}$ in distribution (Eq.~\ref{eq:our_method_collapse}). One-hot SFT thus appears as the degenerate zero-temperature limit of the GIFT target, rather than as a separate construction.

\section{Two-Stage Decomposition Identity}
\label{app:two_stage_proof}

We establish the additive identity stated in Eq.~\ref{eq:decomposition} of the main text. Adopting the notation of Appendix~\ref{app:token_dv} and writing $\pi_{\text{sft}}^*$ for the token-level target in Eq.~\ref{eq:app_token_gibbs}, we show that for any $0 \leq \beta \leq \bar{\eta}$ and any $q \in \Delta(\mathcal{V})$,
\begin{equation}
\label{eq:app_decomp}
\mathcal{F}_{\bar{\eta}}(q) = \mathcal{F}_\beta(\pi_{\text{sft}}^*) + \mathcal{F}^{(2)}_{\bar{\eta} - \beta}\!\big(q;\, \pi_{\text{sft}}^*\big),
\end{equation}
where $\mathcal{F}_\beta(q) = \beta \mathbb{E}_q[r] - D_{\text{KL}}(q \| \pi_{\text{base}})$ and $\mathcal{F}^{(2)}_{\bar{\eta} - \beta}(q; \pi_{\text{sft}}^*) = (\bar{\eta} - \beta) \mathbb{E}_q[r] - D_{\text{KL}}(q \| \pi_{\text{sft}}^*)$.

\paragraph{Algebraic derivation.} Using the closed form $\pi_{\text{sft}}^*(y) = \pi_{\text{base}}(y) \mathrm{e}^{\beta r(y)} / Z_t(\beta)$, we expand the residual KL term:
\begin{equation*}
\begin{aligned}
D_{\text{KL}}(q \| \pi_{\text{sft}}^*) &= \mathbb{E}_q\!\left[ \log \frac{q(y)}{\pi_{\text{sft}}^*(y)} \right] \\
&= \mathbb{E}_q[\log q(y)] - \mathbb{E}_q[\log \pi_{\text{base}}(y)] \\
& \quad - \beta \, \mathbb{E}_q[r(y)] + \log Z_t(\beta) \\
&= D_{\text{KL}}(q \| \pi_{\text{base}}) - \beta \, \mathbb{E}_q[r] + \log Z_t(\beta).
\end{aligned}
\end{equation*}
Substituting into $\mathcal{F}^{(2)}_{\bar{\eta} - \beta}(q; \pi_{\text{sft}}^*)$ and rearranging,
\begin{equation*}
\begin{aligned}
\mathcal{F}^{(2)}_{\bar{\eta} - \beta}(q; \pi_{\text{sft}}^*) 
&= (\bar{\eta} - \beta) \, \mathbb{E}_q[r] - D_{\text{KL}}(q \| \pi_{\text{sft}}^*) \\
&= \bar{\eta} \, \mathbb{E}_q[r] - D_{\text{KL}}(q \| \pi_{\text{base}}) - \log Z_t(\beta) \\
&= \mathcal{F}_{\bar{\eta}}(q) - \log Z_t(\beta).
\end{aligned}
\end{equation*}
By the Donsker--Varadhan duality (Appendix~\ref{app:token_dv}), $\log Z_t(\beta) = \mathcal{F}_\beta(\pi_{\text{sft}}^*)$, since the supremum of $\mathcal{F}_\beta$ is attained at $\pi_{\text{sft}}^*$. Substituting back yields Eq.~\ref{eq:app_decomp}.

\paragraph{Two-stage suprema match.} Maximizing both sides of Eq.~\ref{eq:app_decomp} over $q$, the left side gives $\log Z_t(\bar{\eta})$ by Donsker--Varadhan applied at inverse temperature $\bar{\eta}$. On the right side, $\mathcal{F}_\beta(\pi_{\text{sft}}^*)$ is constant in $q$ and equals $\log Z_t(\beta)$, while $\sup_q \mathcal{F}^{(2)}_{\bar{\eta}-\beta}(q; \pi_{\text{sft}}^*) = \log \mathbb{E}_{\pi_{\text{sft}}^*}[\mathrm{e}^{(\bar{\eta} - \beta) r}]$ by the same duality applied to the residual problem with reference $\pi_{\text{sft}}^*$. A direct computation using $\pi_{\text{sft}}^*(y) \mathrm{e}^{(\bar{\eta} - \beta) r(y)} = \pi_{\text{base}}(y) \mathrm{e}^{\bar{\eta}\, r(y)} / Z_t(\beta)$ gives
\begin{equation*}
\log \mathbb{E}_{\pi_{\text{sft}}^*}\!\big[\mathrm{e}^{(\bar{\eta} - \beta) r}\big] = \log Z_t(\bar{\eta}) - \log Z_t(\beta),
\end{equation*}
so the two suprema agree: the SFT-then-RL pipeline (stage one maximizing $\mathcal{F}_\beta$, stage two maximizing $\mathcal{F}^{(2)}_{\bar{\eta} - \beta}$ with reference $\pi_{\text{sft}}^*$) attains the same value as the single-stage problem at the aggregate inverse temperature $\bar{\eta}$.

\paragraph{Structural correspondence with KL-regularized RL.} Dividing the residual term in Eq.~\ref{eq:residual} by $(\bar{\eta} - \beta) > 0$ does not change its argmax over $q$, so $\mathcal{F}^{(2)}_{\bar{\eta} - \beta}(q; \pi_{\text{sft}}^*)$ shares its maximizer with the per-token KL-regularized objective
\begin{equation*}
\mathbb{E}_q[r] - \tfrac{1}{\bar{\eta}-\beta}\, D_{\text{KL}}\!\big(q \,\big\|\, \pi_{\text{sft}}^*\big).
\end{equation*}
Identifying this with the standard form of Appendix~\ref{appendix:derivation} gives the structural correspondence $\lambda = 1/(\bar{\eta} - \beta)$, so $\bar{\eta} - \beta$ plays the role of the inverse temperature contributed by the RL stage.

We emphasize that this is a structural correspondence in form: the decomposition does not require the trajectory-level RL reward $R(x,y)$ in Eq.~\ref{eq:RL_obj} to coincide with the per-token oracle indicator $r$ used here. Any token-level credit assignment of $R$ (e.g., per-token GRPO advantages) realizes the residual stage in the same form, with the empirical analog of $\bar{\eta} - \beta$ shaped by the actual KL coefficient and the implicit trust-region effects of clipping and finite-step updates. We accordingly treat $\bar{\eta}$ as a structural surrogate that organizes the two stages, and select $\beta$ as a hyperparameter rather than tying it to a closed-form expression of $\lambda$ (Section~\ref{sec:beta_analysis}).

\paragraph{Degenerate limit $\beta \to \infty$.} As $\beta \to \infty$, $\pi_{\text{sft}}^* \to \delta_{y^*_t}$ (Appendix~\ref{app:token_dv}). The residual term $D_{\text{KL}}(q \| \pi_{\text{sft}}^*)$ in Eq.~\ref{eq:residual} then diverges for every $q$ that places nonzero mass off $y^*_t$, so the residual feasible region on which $\mathcal{F}^{(2)}_{\bar{\eta} - \beta}$ is finite contracts strictly to $\{\delta_{y^*_t}\}$. The residual stage is then unable to refine the policy in any meaningful direction, recovering the prior collapse mode discussed in Section~\ref{sec:two_stage_bridge}. For finite $\beta$, $D_{\text{KL}}(q \| \pi_{\text{sft}}^*)$ remains finite for every $q$ supported on $\mathcal{V}$, leaving the residual stage well-posed and providing room for the RL stage to absorb the remaining inverse temperature $\bar{\eta} - \beta$.

\section{Training Dynamics.} 
\begin{figure}[htbp]
    \centering
    \includegraphics[width=1.0\linewidth]{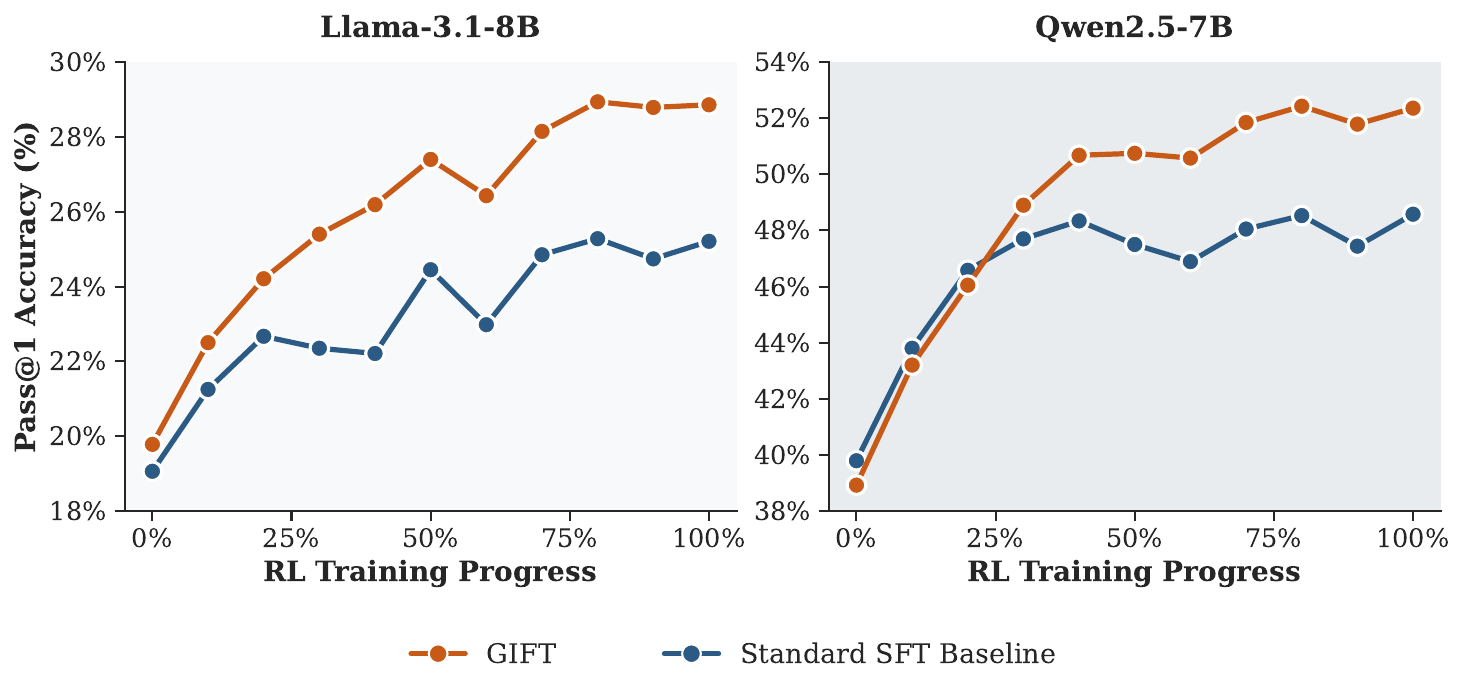}
    \caption{\textbf{Average pass@1 accuracy across the RL training progress.}}
    \label{fig:training_dynamics}
\end{figure}

We report the accuracy on mathematical reasoning benchmarks during the RL training in Figure \ref{fig:training_dynamics}. GIFT initialization demonstrates superior sample efficiency and asymptotic performance compared to standard SFT initialization. On Llama-3.1-8B, GIFT maintains a consistent performance advantage over the SFT baseline throughout the entire optimization process. Meanwhile, on Qwen2.5-7B, we observe a crossover phenomenon: although GIFT begins with slightly lower performance, it rapidly overtakes the SFT baseline within the first 25\% steps. This trajectory indicates that GIFT effectively avoids the premature convergence often induced by rigid supervision, trading off negligible initial exploitation for substantially greater exploration potential and superior long-term gains.

\section{Generalizability on Open-ended Generation Tasks}
\label{app:open-ended}
While our primary evaluation focuses on verifiable mathematics and reasoning, we extend our experiments to the \textbf{MT-Bench} benchmark~\cite{Bai_2024} to verify whether the token-level GIFT target maintains generalizability in unconstrained, open-ended conversational settings. As shown in Table~\ref{tab:mtbench_appendix}, GIFT consistently achieves the highest overall scores (\textit{Total}) across both Qwen2.5-7B and Llama-3.1-8B models. Specifically, it excels in categories such as STEM and Extraction, while avoiding the severe performance degradation often observed in standard SFT variants (e.g., SFT+KD or Label Smoothing) when faced with diverse user prompts. These results demonstrate that GIFT successfully preserves the generative priors of the base model, ensuring robust open-ended dialogue capabilities while simultaneously achieving peak reasoning potential.

\section{Detailed Performance Prior to RL Stage}
\label{sec:detailed_results}

Table \ref{tab:detailed_pass_at_k} details the pass@$k$ performance across individual mathematical reasoning benchmarks. We observe a clear trade-off between strict mode-seeking and exploration: while standard SFT remains highly competitive at $k=1$ due to its rigid optimization of the target trajectories, it exhibits severe exploration bottlenecks at larger $k$ values. In contrast, GIFT exhibits robust scaling behavior. As the sample budget scales to $k=8$, GIFT consistently overtakes the baseline, confirming that a finite-temperature initialization successfully maintains the diversity necessary for broader exploration.

\begin{table*}[t]
\centering
\small
\caption{Detailed pass@$k$ performance (\%) on individual mathematical reasoning benchmarks prior to RL.}
\label{tab:detailed_pass_at_k}
\begin{tabular}{llcccccccccc}
\toprule
\multirow{2}{*}{\textbf{Benchmark}} & \multirow{2}{*}{\textbf{Strategy}} & \multicolumn{4}{c}{\textbf{Llama-3.1-8B}} & & \multicolumn{4}{c}{\textbf{Qwen2.5-7B}} \\
\cmidrule{3-6} \cmidrule{8-11}
& & \textbf{$k=1$} & \textbf{$k=2$} & \textbf{$k=4$} & \textbf{$k=8$} & & \textbf{$k=1$} & \textbf{$k=2$} & \textbf{$k=4$} & \textbf{$k=8$} \\
\midrule
\multirow{2}{*}{\textbf{GSM8K}} 
 & SFT & \textbf{73.84} & 86.20 & \textbf{92.97} & 96.11 & & \textbf{90.14} & 94.58 & 96.35 & 97.35 \\
 & \textbf{GIFT} & 73.36 & \textbf{86.65} & 92.60 & \textbf{96.35} & & 89.94 & \textbf{94.71} & \textbf{96.73} & \textbf{97.80} \\
\midrule
\multirow{2}{*}{\textbf{AIME 24}} 
 & SFT & \textbf{0.97} & \textbf{1.94} & \textbf{3.89} & 7.78 & & \textbf{11.25} & 17.02 & 21.86 & 23.33 \\
 & \textbf{GIFT} & 0.83 & \textbf{1.94} & 3.33 & \textbf{8.33} & & 10.00 & \textbf{19.88} & \textbf{24.29} & \textbf{30.00} \\
\midrule
\multirow{2}{*}{\textbf{AIME 25}} 
 & SFT & \textbf{0.55} & \textbf{1.11} & 2.22 & 4.44 & & \textbf{11.25} & 14.88 & 17.62 & 20.00 \\
 & \textbf{GIFT} & 0.42 & \textbf{1.11} & \textbf{2.50} & \textbf{5.00} & & 10.00 & \textbf{18.21} & \textbf{24.71} & \textbf{33.33} \\
\midrule
\multirow{2}{*}{\textbf{MATH-500}} 
 & SFT & \textbf{35.94} & 48.15 & 59.58 & 68.67 & & 73.92 & 82.14 & 87.77 & 91.40 \\
 & \textbf{GIFT} & 35.80 & \textbf{48.42} & \textbf{60.06} & \textbf{69.80} & & \textbf{74.30} & \textbf{82.86} & \textbf{88.19} & \textbf{91.60} \\
\bottomrule
\end{tabular}
\end{table*}

\begin{table*}[htbp]
\centering
\scriptsize
\caption{Comprehensive performance on the MT-Bench benchmark across various conversational and reasoning categories.}
\label{tab:mtbench_appendix}
\resizebox{\textwidth}{!}{
\begin{tabular}{lccccccccc}
\toprule
\textbf{Strategy} & \textbf{Total} & \textbf{Writing} & \textbf{Roleplay} & \textbf{Reasoning} & \textbf{Math} & \textbf{Coding} & \textbf{Extract} & \textbf{STEM} & \textbf{Humanities} \\
\midrule
\multicolumn{10}{c}{\textit{\textbf{Qwen2.5-7B}}} \\
\midrule
SFT & 5.79 & 4.0 & \textbf{7.7} & 4.1 & 5.3 & 5.3 & 5.0 & 6.7 & 8.2 \\
SFT + Entropy & 6.12 & 4.4 & \textbf{7.7} & \textbf{6.7} & 4.7 & 5.0 & 5.7 & 7.0 & 7.8 \\
SFT + Label Smoothing & 6.00 & 4.7 & 6.6 & 4.0 & \textbf{6.4} & 5.2 & 5.6 & 6.8 & \textbf{8.7} \\
SFT + KD & 6.12 & \textbf{5.5} & 7.2 & 4.5 & 5.8 & \textbf{5.7} & 5.6 & 7.1 & 7.6 \\
DFT & 5.60 & 4.1 & 5.8 & 4.0 & 6.3 & 5.1 & 5.8 & 5.9 & 7.8 \\
ASFT & 5.85 & 3.7 & 6.9 & 4.3 & 6.1 & 5.6 & 5.6 & 6.5 & 8.1 \\
PSFT & 5.64 & 3.7 & 6.7 & 4.1 & 4.1 & \textbf{5.7} & 5.2 & 7.0 & 8.6 \\
\textbf{GIFT} & \textbf{6.34} & 5.2 & 6.9 & 4.8 & \textbf{6.4} & 5.2 & \textbf{6.5} & \textbf{7.6} & 8.1 \\
\midrule
\multicolumn{10}{c}{\textit{\textbf{Llama3.1-8B}}} \\
\midrule
SFT & 4.46 & 4.9 & 5.4 & 3.5 & 2.5 & 3.4 & 3.8 & 4.7 & \textbf{7.5} \\
SFT + Entropy & 3.94 & 4.7 & 4.3 & 2.8 & 2.4 & 2.5 & 3.7 & 4.6 & 6.5 \\
SFT + Label Smoothing & 3.99 & 5.0 & 4.5 & 3.0 & 2.8 & 2.1 & 4.5 & 3.8 & 6.2 \\
SFT + KD & 3.30 & 2.8 & 3.3 & 2.2 & 4.0 & 3.3 & 3.0 & 4.1 & 3.7 \\
DFT & 5.05 & 5.0 & \textbf{5.9} & 3.5 & 4.2 & 3.6 & \textbf{5.5} & 5.6 & 7.1 \\
ASFT & 4.89 & 5.2 & 5.6 & 3.8 & 3.6 & \textbf{4.4} & 5.2 & 4.5 & 6.8 \\
PSFT & 4.19 & 5.3 & 4.4 & 3.0 & 3.7 & 2.3 & 2.7 & 5.6 & 6.5 \\
\textbf{GIFT} & \textbf{5.17} & \textbf{6.4} & \textbf{5.9} & \textbf{4.3} & 4.1 & 3.3 & 4.2 & \textbf{6.0} & 7.2 \\
\bottomrule
\end{tabular}
}
\end{table*}

\section{Hyperparameter Sweep for Baselines}
\label{app:regularization_sweep}

GIFT belongs to the broader family of base-anchored soft-target objectives, but differs in target geometry. Knowledge distillation and label smoothing form convex mixtures $q = (1-\alpha)\mathbb{I}[y=y_t^*] + \alpha p$ with $p$ a base or uniform distribution, whereas GIFT applies an exponential tilt $q \propto \pi_{\text{base}}(y)\,\mathrm{e}^{\beta \mathbb{I}[y=y_t^*]}$ that preserves the base relative probabilities among non-oracle tokens while multiplying the oracle-token odds by $\mathrm{e}^{\beta}$. This form is matched to the Gibbs solution of the KL-regularized surrogate, and the sweep below confirms that the gains arise from this geometry rather than generic target softening.

To isolate the structural advantages of GIFT and address the concern that its gains might simply stem from generic target softening, we conduct a comprehensive hyperparameter sweep for Knowledge Distillation (KD) and Label Smoothing (LS) on both Qwen2.5-7B and Llama-3.1-8B. Formally, these baselines modify the standard one-hot target into the following convex mixtures:
\begin{align}
q^{\text{KD}}(y) &= (1 - \alpha) \mathbb{I}[y = y^*_t] + \alpha \pi_{\text{base}}(y), \\
q^{\text{LS}}(y) &= (1 - \mu) \mathbb{I}[y = y^*_t] + \frac{\mu}{|\mathcal{V}|},
\end{align}
where $\mathcal{V}$ is the vocabulary. We vary the distillation weight $\alpha \in \{0.01, 0.03, 0.05, 0.1\}$ and the smoothing parameter $\mu \in \{0.01, 0.03, 0.05, 0.1\}$. The results are summarized in Table~\ref{tab:sweep_baselines}.

\begin{table*}[h]
\scriptsize
\centering
\caption{Hyperparameter sweep for Knowledge Distillation (KD) and Label Smoothing (LS) prior to RL. Standard SFT and GIFT are included as reference bounds. The best result among the swept variants is underlined, while the overall best is bolded.}
\label{tab:sweep_baselines}
\resizebox{\textwidth}{!}{
\begin{tabular}{llccccccc}
\toprule
\textbf{Method} & \textbf{Hyperparameter} & \textbf{GSM8K} & \textbf{AIME24} & \textbf{AIME25} & \textbf{MATH500} & \textbf{HE}$^{+}$ & \textbf{MBPP}$^{+}$ & \textbf{Average} \\
\midrule
\multicolumn{9}{c}{\textit{\textbf{Qwen2.5-7B}}} \\
\midrule
Standard SFT & - & 91.79 & 13.33 & 14.44 & 78.67 & 72.00 & 60.10 & 55.06 \\
\cmidrule{1-9}
\multirow{4}{*}{SFT + KD ($\alpha$)} 
 & $\alpha = 0.01$ & \underline{91.85} & \underline{13.33} & \underline{15.56} & \underline{78.83} & 72.60 & 60.50 & \underline{55.45} \\
 & $\alpha = 0.03$ & 91.62 & 13.33 & 13.33 & 78.37 & 73.00 & 61.20 & 55.14 \\
 & $\alpha = 0.05$ & 91.55 & 10.00 & 11.11 & 78.14 & 73.20 & 61.80 & 54.30 \\
 & $\alpha = 0.10$ & 91.51 & 10.00 & 10.00 & 78.00 & \underline{73.80} & \underline{62.40} & 54.29 \\
\cmidrule{1-9}
\multirow{4}{*}{SFT + LS ($\mu$)} 
 & $\mu = 0.01$ & \underline{90.14} & \underline{13.33} & \underline{10.00} & \underline{73.60} & \underline{72.00} & \underline{59.00} & \underline{53.01} \\
 & $\mu = 0.03$ & 88.47 & 10.00 & 8.89 & 71.23 & 71.50 & 58.40 & 51.42 \\
 & $\mu = 0.05$ & 86.18 & 6.67 & 8.89 & 68.52 & 70.80 & 57.20 & 49.71 \\
 & $\mu = 0.10$ & 82.15 & 3.33 & 6.67 & 62.37 & 69.50 & 55.80 & 46.64 \\
\cmidrule{1-9}
\textbf{GIFT (Ours)} & - & \textbf{92.06} & \textbf{23.33} & \textbf{17.78} & \textbf{82.00} & \textbf{76.20} & \textbf{65.90} & \textbf{59.55} \\
\midrule
\multicolumn{9}{c}{\textit{\textbf{Llama-3.1-8B}}} \\
\midrule
Standard SFT & - & 73.39 & 0.00 & 3.33 & 39.20 & 43.30 & 45.50 & 34.12 \\
\cmidrule{1-9}
\multirow{4}{*}{SFT + KD ($\alpha$)} 
 & $\alpha = 0.01$ & \underline{73.48} & 0.00 & \underline{3.33} & \underline{39.52} & \underline{43.00} & 45.80 & \underline{34.19} \\
 & $\alpha = 0.03$ & 70.12 & \underline{3.33} & 0.00 & 35.18 & 41.50 & 47.10 & 32.87 \\
 & $\alpha = 0.05$ & 64.76 & 3.33 & 0.00 & 30.14 & 39.40 & 48.20 & 30.97 \\
 & $\alpha = 0.10$ & 59.28 & 3.33 & 0.00 & 25.00 & 37.80 & \underline{49.50} & 29.15 \\
\cmidrule{1-9}
\multirow{4}{*}{SFT + LS ($\mu$)} 
 & $\mu = 0.01$ & \underline{67.90} & \underline{0.00} & \underline{0.00} & \underline{33.40} & \underline{36.60} & \underline{47.10} & \underline{30.83} \\
 & $\mu = 0.03$ & 64.23 & 0.00 & 0.00 & 30.08 & 35.20 & 45.50 & 29.17 \\
 & $\mu = 0.05$ & 58.54 & 0.00 & 0.00 & 26.77 & 33.80 & 43.20 & 27.05 \\
 & $\mu = 0.10$ & 49.26 & 0.00 & 0.00 & 20.46 & 31.00 & 40.50 & 23.54 \\
\cmidrule{1-9}
\textbf{GIFT (Ours)} & -- & \textbf{81.33} & \textbf{6.67} & \textbf{6.67} & \textbf{44.45} & \textbf{44.50} & \textbf{50.80} & \textbf{39.07} \\
\bottomrule
\end{tabular}
}
\end{table*}
\section{Implementation Details}
\label{app:details}

In this section, we provide the detailed experimental configuration to facilitate reproducibility. All experiments are conducted using the VeRL framework~\cite{sheng2024hybridflow} on a cluster of 8$\times$NVIDIA H200 GPUs.

\subsection{Supervised Fine-Tuning}
\label{app:sft_details}

\noindent \textbf{Training Configuration.} 
We optimize the base models using the AdamW optimizer with $\beta_1=0.9$, $\beta_2=0.95$, and a weight decay of $0.01$. The global batch size is set to 128. We employ a constant learning rate of $1 \times 10^{-5}$ with no warmup steps. To accommodate long chain-of-thought reasoning, we set the maximum sequence length to 8,192 tokens. Following standard practices for generative fine-tuning, we apply right-side truncation for input sequences. For the explicit SFT ablation variants, we set the regularization coefficient to $\mu=0.01$ for entropy regularization. For all other baseline methods, we strictly adopt the default hyperparameter settings recommended in their respective original papers to ensure a fair comparison. To initialize the subsequent RL stage, we uniformly select the SFT checkpoint from the 6th epoch across all backbones. As analyzed in Section 4.3 (Figure~\ref{fig:beta}), Llama-3.1-8B reaches its lowest validation loss at this epoch before overfitting. For Qwen2.5-7B, the performance differences after epoch 6 are marginal. Thus, we uniformly adopt the 6th epoch to establish a standardized setting and avoid post-hoc selection bias based on RL downstream dynamics. Finally, to ensure a fair comparison, the direct SFT baseline is trained on a combined dataset comprising both SFT and RL data to match the total data size configuration.

\subsection{Reinforcement Learning}
Following SFT, we apply the Group Relative Policy Optimization (GRPO) algorithm~\cite{shao2024deepseekmath} to refine the policy. During the exploration phase, we sample $G=8$ outputs per prompt with a temperature of $T=1.0$ to encourage diversity. The policy is optimized using a constant learning rate of $1 \times 10^{-6}$ and a global batch size of 128 (mini-batch size of 64). We set the clip ratio to 0.2 and disable advantage normalization (\texttt{norm\_adv\_by\_std=False}). To prevent reward hacking, we restrict training to 1 epoch.

\subsection{Evaluation and Reward Mechanism}
\noindent \textbf{Evaluation.} We report the average pass@1 accuracy across four runs for all benchmarks. We merge FSDP checkpoints and utilize vLLM for high-throughput inference with a temperature of $T=0.6$ and a max token limit of 8,192. A single $\beta$ value is applied uniformly across all evaluated benchmarks without per-task tuning.

\noindent \textbf{Reward Function.} We employ a cascaded binary reward function that sequentially utilizes \texttt{OpenMathInstruct} extraction rules and the \texttt{MathVerify} parser. To ensure stability, the verification process is guarded by a 10-second thread-based timeout, assigning a reward of 1.0 if either method validates the solution against the ground truth.

\section{Use of AI Assistants}
We utilized AI assistants exclusively for grammatical polishing and LaTeX formatting assistance. 